\documentclass[11pt,a4paper]{article}
\usepackage[hyperref]{emnlp2020}
\usepackage{times}
\usepackage{latexsym}
\usepackage{amsmath}
\usepackage{amsfonts}
\usepackage{graphicx,subfigure}
\usepackage{subfigure}
\usepackage{booktabs}
\usepackage{color, soul}
\usepackage{tablefootnote}
\usepackage{threeparttable}
\usepackage{wrapfig}
\usepackage{dirtytalk}

\usepackage{algorithm} 
\usepackage[noend]{algpseudocode} 

\floatname{algorithm}{Procedure}

\usepackage{float}
\usepackage{hyperref}

\usepackage{microtype}

\aclfinalcopy 

\title{Multimodal Routing: Improving Local and Global Interpretability of Multimodal Language Analysis}
\author{%
  Yao-Hung Hubert Tsai\thanks{\ indicates equal contribution. Code is available at \url{https://github.com/martinmamql/multimodal_routing}.}\ , Martin Q. Ma\footnotemark[1]\ , Muqiao Yang\footnotemark[1]\ , \\ {\bf Ruslan Salakhutdinov, Louis-Philippe Morency}\\
 Carnegie Mellon University, Pittsburgh, PA, USA \\
 \texttt{\{yaohungt, qianlim, muqiaoy, rsalakhu, lmorency\}@cs.cmu.edu}
}

\date{}

\begin{document}
\maketitle
\begin{abstract}
The human language can be expressed through multiple sources of information known as modalities, including tones of voice, facial gestures, and spoken language. Recent multimodal learning with strong performances on human-centric tasks such as sentiment analysis and emotion recognition are often black-box, with very limited interpretability. In this paper we propose \textit{Multimodal Routing}, which dynamically adjusts weights between input modalities and output representations differently for each input sample. Multimodal routing can identify relative importance of both individual modalities and cross-modality features. Moreover, the weight assignment by routing allows us to interpret modality-prediction relationships not only globally (i.e. general trends over the whole dataset), but also locally for each single input sample, meanwhile keeping competitive performance compared to state-of-the-art methods. 

\end{abstract}

\section{Introduction}
The human language contains multimodal cues, including textual (e.g., spoken or written words), visual (e.g., body gestures), and acoustic (e.g., voice tones) modalities. It acts as a medium for human communication and has been advanced in areas spanning affect recognition~\cite{busso2008iemocap}, media description~\cite{lin2014microsoft}, and multimedia information retrieval~\cite{abu2016youtube}. Modeling multimodal sources requires to understand the relative importance of not only each single modality (defined as {\em unimodal explanatory features}) but also the interactions (defined as {\em bimodal} or {\em trimodal explanatory features})~\cite{buchel1998multimodal}. Recent work ~\cite{liu2018efficient, williams2018recognizing, ortega2019multimodal} proposed methods to fuse information across modalities and yielded superior performance, but these models are often black-box with very limited interpretability. 
\begin{figure}[H]
    \centering
    \includegraphics[scale=0.6]{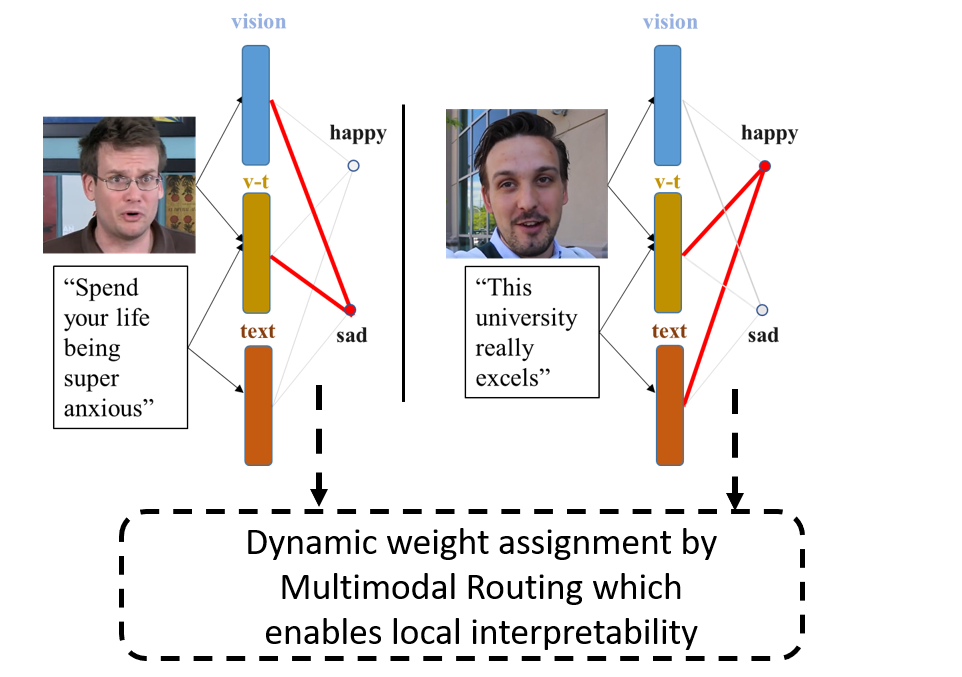}
    \caption{An example of Multimodal Routing, where the weights between visual, textual, and visual-textual explanatory features and concepts of emotions (happy and sad) are dynamically adjusted given every input sample. The model associates vision and $v$-$t$ features to sad concept in the left sample, and $v$-$t$ and text features to happy concept in the right example, showing local weights interpretation upon different input features.}
    \label{fig:example}
\end{figure}
\begin{figure*}[h!]
    \centering
    \includegraphics[width=\textwidth]{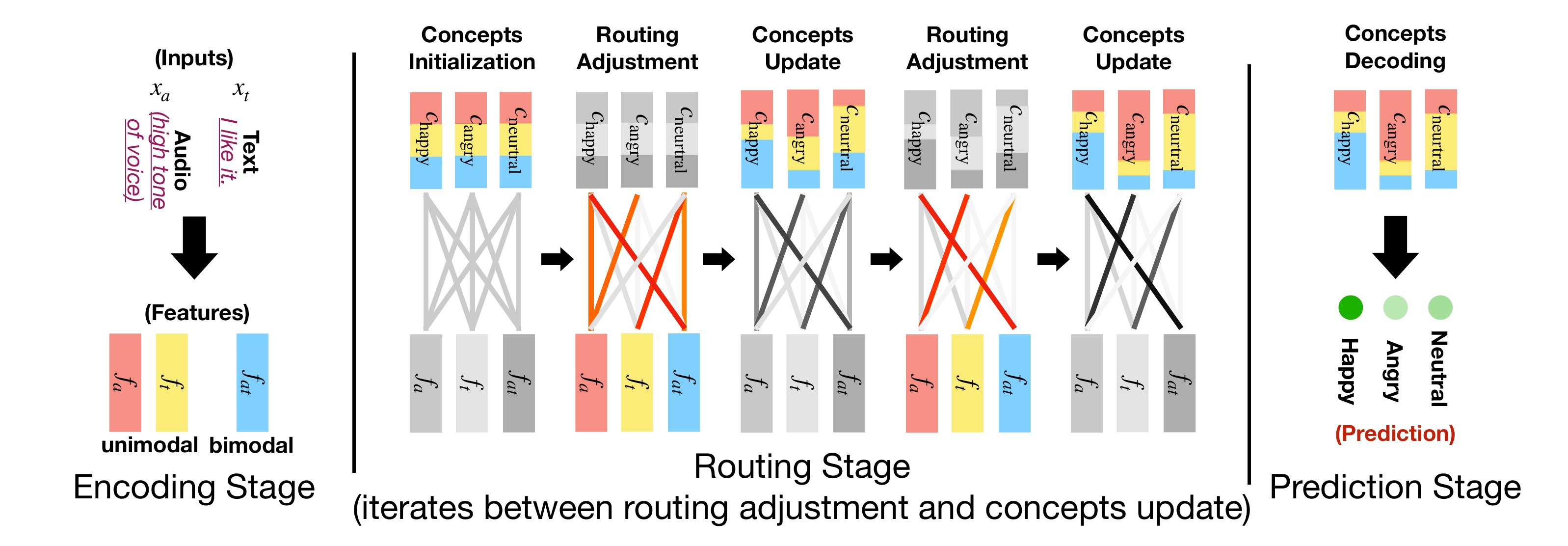}
    \vspace{-6mm}
    \caption{Overview of Multimodal Routing, which contains encoding, routing, and prediction stages. We consider only two input modalities in this example.
    The encoding stage computes unimodal and bimodal explanatory features with the inputs from different modalities. The routing stage iteratively performs {\em concepts update} and {\em routing adjustment}. The prediction stage decodes the concepts to the model's prediction. The routing associates the text and the visual-text features with negative sentiment in the left example, and the vision and the visual-text features with positive sentiment in the right example \textit{before making predictions}. }
    \vspace{-1mm}
    \label{fig:illus}
\end{figure*}

Interpretability matters. It allows us to identify crucial explanatory features for predictions. Such interpretability knowledge could be used to provide insights into multimodal learning, improve the model design, or debug a dataset. This inerpretability is useful at two levels: the global and the local level. The global interpretation reflects the general (averaged) trends of explanatory feature importance over the whole dataset. The local interpretation is arguably harder but can give a high-resolution insight of feature importance specifically depending on each \textit{individual} samples during training and inference. These two levels of interpretability should provide us an understanding of unimodal, bimodal and trimodal explanatory features.

In this paper we address both local and global interpretability of unimodal, bimodal and trimodal explanatory featuress by presenting \textit{Multimodal Routing}. In human multimodal language, such routing dynamically changes weights between modalities and output labels for each sample as shown in Fig.~\ref{fig:example}. The most significant contribution of Multimodal Routing is its ability to establish local weights dynamically for each input sample between modality features and the labels during training and inference, thus providing local interpretation for each sample. 

Our experiments focus on two tasks of sentiment analysis and emotion recognition tasks using two benchmark multimodal language datasets, IEMOCAP~\cite{busso2008iemocap} and CMU-MOSEI~\cite{zadeh2018multimodal}. We first study how our model compares with the state-of-the-art methods on these tasks. More importantly we provide local interpretation by qualitatively analyzing adjusted local weights for each sample. Then we also analyze the global interpretation using statistical techniques to reveal crucial features for prediction on average. Such interpretation of different resolutions strengthens our understanding of multimodal language learning.

\section{Related Work}




Multimodal language learning is based on the fact that human integrates multiple sources such as acoustic, textual, and visual information to learn language~\cite{mcgurk1976hearing, ngiam2011multimodal, baltruvsaitis2018multimodal}. Recent advances in modeling multimodal language using deep neural networks are not interpretable ~\cite{wang2019words,tsai2019multimodal}. Linear method like the Generalized Additive Models (GAMs)~\cite{hastie2017generalized} do not offer local interpretability. Even though we could use post hoc (interpret predictions given an arbitrary model) methods such as LIME~\cite{ribeiro2016should}, SHAP~\cite{lundberg2017unified}, and L2X~\cite{chen2018learning} to interpret these black-box models, these interpretation methods are designed to detect the contributions only from unimodal features but not bimodal or trimodal explanatory features. It is shown that in human communication,  modality interactions are more important than individual modalities~\cite{engle1998not}.

Two recent methods, Graph-MFN~\cite{zadeh2018multimodal} and Multimodal Factorized Model (MFM)~\cite{tsai2019learning}, attempted to interpret relationships between modality interactions and learning for human language. Nonetheless, Graph-MFN did not separate the contributions among unimodal and multimodal explanatory features, and MFM only provided the analysis on trimodal interaction feature. Both of them cannot interpret how both single modality and modality interactions contribute to final prediction at the same time.




Our method is inspired and related to Capsule Networks~\cite{sabour2017dynamic,hinton2018matrix}, which performs routing between layers of capsules. Each capsule is a group of neurons that encapsulates spatial information as well as the probability of an object being present. On the other hand, our method performs routing between multimodal features (i.e., unimodal, bimodal, and trimodal explanatory features) and concepts of the model's decision.

\section{Method}
The proposed Multimodal Routing contains three stages shown in Fig.~\ref{fig:illus}: encoding, routing, and prediction. The encoding stage encodes raw inputs (speech, text, and visual data) to unimodal, bimodal, and trimodal features. The routing stage contains a routing mechanism~\cite{sabour2017dynamic,hinton2018matrix}, which 1) updates some hidden representations and 2) adjusts local weights between each feature and each hidden representation by pairwise similarity. Following previous work~\cite{mao2019neuro}, we call the hidden representations ``concepts'', and each of them is associated to specific a prediction label (in our case sentiment or an emotion). Finally, the prediction stage takes the inferred concepts to perform model prediction. 


\subsection{Multimodal Routing}
\label{subsec:method}
We use $v$(isual), $a$(coustic), and $t$(ext) to denote the three commonly considered modalities in human multimodal language. 
Let $x = \{x_a, x_v, x_t\}$ represent the multimodal input. 
$x_a \in \mathbb{R}^{T_a \times d_a}$ is an audio stream with time length $T_a$ and feature dimension $d_a$ (at each time step). Similarly, $x_v \in \mathbb{R}^{T_v \times d_v}$ is the visual stream and $x_t \in \mathbb{R}^{T_t \times d_t}$ is the text stream. In our paper, we consider multiclass or multilabel prediction tasks for the multimodal language modeling, in which we use $y\in \mathbb{R}^J$
to denote the ground truth label with $J$ being the number of classes or labels, and $\hat{y}$ to represent the model's prediction. Our goal is to find the relative importance of the contributions from unimodal (e.g., $x_a$ itself), bimodal (e.g., the interaction between $x_a$ and $x_v$), and trimodal features (e.g., the interaction between $x_a$, $x_v$, and $x_t$) to the model prediction $\hat{y}$.

\vspace{1.5mm}
\hspace{-4mm}{\bf Encoding Stage.} The encoding stage encodes multimodal inputs $\{x_a, x_v, x_t\}$ into explanatory features. We use $f_{i} \in \mathbb{R}^{d_f}$ to denote the features with $i\in\{a,v,t\}$ being unimodal, $i\in\{av,vt,ta\}$ being bimodal, and $i\in \{avt\}$ being trimodal interactions. $d_f$ is the dimension of the feature. To be specific, $f_a = F_{a}(x_a; \theta)$, $f_{av} = F_{av}(x_a, x_v; \theta)$, and $f_{avt} = F_{avt}(x_a, x_v, x_t; \theta)$ with $\theta$ as the parameters of the encoding functions and $F$ as the encoding functions. Multimodal Transformer (MulT)~\cite{tsai2019multimodal} is adopted as the design of the encoding functions $F_i$. Here the trimodal function $F_{avt}$ encodes sequences from three modalities into a unified representation, 
$F_{av}$ encodes acoustic and visual modalities, and $F_{a}$ encodes acoustic input. Next, $p_i \in [0,1]$ is a scalar representing how each feature $f_i$ is activated in the model. Similar to $f_i$, we also use MulT to encode $p_i$ from the input $x_i$. That is, $p_a = P_{a}(x_a; \theta')$, $p_{av} = P_{av}(x_a, x_v, \theta')$, and $p_{avt} = P_{avt}(x_a, x_v, x_t, \theta')$ with $\theta'$ as the parameters of MulT and $P_i$ as corresponding encoding functions (details in the Supplementary). 


\vspace{1.5mm}
\hspace{-4mm}{\bf Routing Stage.} The goal of routing is to infer interpretable hidden representations (termed here as ``concepts'') for each output label. The first step of routing is to initialize the concepts with equal weights, where all explanatory features are as important. Then the core part of routing is an iterative process which will enforce for each explanatory feature to be assigned to only one output representations (a.k.a the ``concepts''; in reality it is a soft assignment) that shows high similarity with a concept. Formally each concept $c_j \in \mathbb{R}^{d_c}$ is represented as a one-dimensional vector of dimension $d_c$. Linear weights $r_{ij}$, which we term \textit{routing coefficient}, are defined between each concept $c_j$ and explanatory factor $f_i$. 

The first half of routing, which we call \textit{routing adjustment}, is about finding new assignment (i.e. the routing coefficients) between the input features and the newly learned concepts by taking a softmax of the dot product over all concepts, thus only the features showing high similarity of a concept (sentiment or an emotion in our case) will be assigned close to $1$, instead of having all features assigned to all concepts. This will help local interpretability because we can always distinguish important explanatory features from non-important ones. The second half of routing, which we call \textit{concept update}, is to update concepts by linearly aggregating the new input features weighted by the routing coefficients, so that it is local to each input samples. 



\vspace{0.5mm}
\hspace{-4mm}\underline{\em - Routing adjustment.} We define the \textit{routing coefficient} $r_{ij} \in [0,1]$ by measuring the similarity~\footnote{We use dot-product as the similarity measurement as in prior work~\cite{sabour2017dynamic}. Note that routing only changes $r_{ij}$, not $W_{ij}$. Another choice can be the probability of a fit under a Gaussian distribution~\cite{hinton2018matrix}.} between $f_i W_{ij}$ and $c_j$:
\begin{equation}
r_{ij} =\frac{\exp(
\left \langle f_i W_{ij}, c_j \right \rangle
)}{\sum_{j'}\exp(
\left \langle f_i W_{ij'}, c_{j'} \right \rangle}
\end{equation}
We note that $r_{ij}$ is normalized over all concepts $c_j$. Hence, it is a coefficient which takes high value only when $f_i$ is in agreement with $c_j$ but not with $c_{j'}$, where $j'\neq j$.

\vspace{0.5mm}
\hspace{-4mm}\underline{\em - Concept update.} After obtaining $p_i$ from encoding stage, we update concepts $c_j$ using weighted average as follows: updates the concepts based on the routing weights by summing input features $f_i$ projected by weight matrix $W_{ij}$ to the space of the j$th$ concept
\begin{equation}
c_j=\sum_i p_i r_{ij} (f_i W_{ij})
\end{equation}
$c_j$ is now essentially a linear aggregation from $(f_i W_{ij})$ with weights $p_i r_{ij}$.

\setlength{\textfloatsep}{5pt}
\begin{algorithm}[t]
\footnotesize
\caption{\footnotesize Multimodal Routing}

\label{algo:routingalg}
\begin{algorithmic}[1]
\Procedure{Routing}{$\{f_i\}$, $\{p_i\}$, $\{W_{ij}\}$}
\State Concepts are initialized with uniform weights
\For{$t$ iterations}
\Statex \quad \quad \quad {\color{black} \it/* Routing Adjustment */}
\State for all feature $i$ and concept $j$: $s_{ij} \gets (f_i W_{ij})^\top  c_{j}$
\State for all feature $i$: $r_{ij} \gets {\rm exp}\big(s_{ij}\big) / \sum_{j'}{\rm exp}\big(s_{i{j'}}\big)$
\Statex \quad \quad \quad  {\color{black} \it/* Concepts Update */}
\State for all concept $j$: $c_{j} \gets \sum_{i}p_i r_{ij} (f_i W_{ij})$

\Return{$\{c_j\}$}


\EndFor
\EndProcedure
\end{algorithmic}
\label{algo:routing}
\end{algorithm}

We summarize the routing procedure in Procedure ~\ref{algo:routing}, which returns concepts ($c_j$) given explanatory features ($f_i$), local weights ($W_{ij}$) and $p_i$. First, we initialize the concepts with uniform weights. Then, we iteratively perform adjustment on routing coefficients ($r_{ij}$) and concept updates. Finally, we return the updated concepts. 

\vspace{1.5mm}
\hspace{-4mm}{\bf Prediction Stage.} The prediction stage takes the inferred concepts to make predictions $\hat{y}$. Here, we apply linear transformations to concept $c_j$ to obtain the logits. Specifically, the $j$th logit is formulated as
\begin{equation}
\begin{split}
{\rm logit}_j & = o_j^\top c_j \\
& = \sum_i p_i r_{ij} o_j^\top (f_{i}W_{ij})
\end{split}
\label{eq:logit}
\end{equation}
where $o_j \in \mathbb{R}^{d_c}$ and is the weight of the linear transformation for the $j$th concept. Then, the \textit{Softmax} (for multi-class task) or \textit{Sigmoid} (for multi-label task) function is applied on the logits to obtain the prediction $\hat{y}$.

\subsection{Interpretability}
\label{subsec:interpret}
In this section, we provide the framework of locally interpreting relative importance of unimodal, bimodal, and trimodal explanatory features to model prediction given different samples, by interpreting the routing coefficients $r_{ij}$, which represents the weight assignment between feature $f_i$ and concept $c_j$. We also provide methods to globally interpret the model across the whole dataset.

\subsubsection{Local Interpretation}

The goal of local interpretation is trying to understand how the importance of modality and modality interaction features change, given different multimodal samples. In eq.~\eqref{eq:logit}, a decision logit considers an addition of the contributions from the unimodal $\{f_{a}, f_v, f_t\}$, bimodal $\{f_{av}, f_{vt}, f_{ta}\}$, and trimodal $f_{avt}$ explanatory features. The particular contribution from the feature $f_i$ to the $j$th concept is represented by $p_i r_{ij} o_j^\top (f_i W_{ij})$. It takes large value when 1) $p_i$ of the feature $f_i$ is large; 2) the agreement $r_{ij}$ is high (the feature $f_i$ is in agreement with concept $c_j$ and is not in agreement with $c_{j'}$, where $j' \neq j$); and 3) the dot product $o_j^\top (f_i W_{ij})$ is large. Intuitively, any of the three scenarios requires high similarity between a modality feature and a concept vector which represents a specific sentiment or emotion. Note that $p_i$, $r_{ij}$ and $f_i$ are the covariates and $o_j$ and $W_{ij}$ are the parameters in the model. Since different input samples yield distinct $p_i$ and $r_{ij}$, we can locally interpret $p_i$ and $r_{ij}$ as the effects of the modality feature $f_i$ contributing to the $j$th logit of the model, which is roughly a confidence of predicting $j$th sentiment or emotion. We will show examples of local interpretations in the Interpretation Analysis section.

\subsubsection{Global Interpretation}

To globally interpret Multimodal Routing, we analyze $\overline{r_{ij}}$, the average values of routing coefficients $r_{ij}$s over the entire dataset. Since eq.~\eqref{eq:logit} considers a linear effect from $f_i$, $p_i$ and $r_{ij}$ to ${\rm logit}_j$, $\overline{r_{ij}}$ represents the {\it average} assignment from feature $f_i$ to the $j$th logit. Instead of reporting the values for $\overline{r_{ij}}$, we provide a statistical interpretation on $\overline{r_{ij}}$ using confidence intervals to provide a range of possible plausible coefficients with probability guarantees. Similar tests on $p_i$ and $p_ir_{ij}$ are provided in Supplementary Materials. Here we choose confidence intervals over p-values because they provide much richer information~\cite{ranstam2012p,du2009confidence}. Suppose we have $n$ data with the corresponding ${\bf r}_{ij} = \{r_{ij, 1}, r_{ij, 2}, \cdots r_{ij, n}\}$. If $n$ is large enough and ${\bf r}_{ij}$ has finite mean and finite variance (it suffices since $r_{ij}\in [0,1]$ is bounded), according to Central Limit Theorem, $\overline{r_{ij}}$ (i.e., mean of ${\bf r}_{ij}$) follows a Normal distribution:
\begin{equation}
\overline{r_{ij}} \sim \mathcal{N} \left( \mu, \frac{s_n^2}{n} \right),
\label{eq:sample_mean_normal}
\end{equation}
where $\mu$ is the true mean of $r_{ij}$ and $s_n^2$ is the sample variance in ${\bf r}_{ij}$. Using eq.~\eqref{eq:sample_mean_normal}, we can provide a confidence interval for $\overline{r_{ij}}$. We follow $95\%$ confidence in our analysis.

\begin{table*}[ht]
\small
\centering
\resizebox{0.93\textwidth}{!}{ 
\begin{tabular}{c|ccc|cccccccc}
\toprule
 & \multicolumn{3}{c|}{CMU-MOSEI Sentiment} & \multicolumn{8}{c}{IEMOCAP Emotion} \\
  Models & \multicolumn{3}{c|}{-} & \multicolumn{2}{c}{Happy} & \multicolumn{2}{c}{Sad} & \multicolumn{2}{c}{Angry} & \multicolumn{2}{c}{Neutral} \\
 & Acc7 & Acc2 & F1 & Acc & F1 & Acc & F1 & Acc & F1 & Acc & F1  \\
\midrule
\multicolumn{12}{c}{Non-Interpretable Methods} \\
\midrule
EF-LSTM & 47.4 & 78.2 & 77.9    & 86.0 & 84.2 & 80.2 & 80.5 & 85.2 & 84.5 & 67.8 & 67.1 \\
LF-LSTM & 48.8 & 80.6 & 80.6    & 85.1 & 86.3 & 78.9 & 81.7 & 84.7 & 83.0 & 67.1 & 67.6 \\
RAVEN~\cite{wang2019words} & 50.0 & 79.1 & 79.5      & 87.3 & 85.8 & 83.4 & 83.1 & 87.3 & 86.7 & 69.7 & 69.3\\
MulT~\cite{tsai2019multimodal} & \textbf{51.8} & \textbf{82.5} & \textbf{82.3}       & \textbf{90.7} & \textbf{88.6} & \textbf{86.7} & \textbf{86.0} & 87.4 & 87.0 & \textbf{72.4} & \textbf{70.7} \\
\midrule
\multicolumn{12}{c}{Interpretable Methods} \\
\midrule
GAM~\cite{hastie2017generalized} & 48.6 & 79.5 & 79.4        & 87.0 & 84.3 & 83.2 & 82.4 & 85.2 & 84.8 & 67.4 & 66.6 \\
Multimodal Routing$^*$  & 50.6 & 81.2 & 81.3   & 85.4 & 81.7 & 84.2 & 83.2 & 83.5 & 83.6 & 67.1 & 66.3\\
Multimodal Routing & \textbf{51.6} & \textbf{81.7} & \textbf{81.8}     & 87.3 & 84.7 & 85.7 & 85.2 & \textbf{87.9} & \textbf{87.7} & 70.4 & \textbf{70.0} \\
\bottomrule
\end{tabular}}
\vspace{-2mm}
\caption{Left: CMU-MOSEI sentiment prediction. Right: IEMOCAP emotion recognition. Multimodal Routing$^*$ denotes our method without iterative routing. Our results are better or close to the state-of-the-art~\cite{tsai2019multimodal}. We make our results bold if it is SOTA or close to SOTA ($\leq 1\%$).}
\label{tab:mosei_senti}
\end{table*}

\begin{table*}[t]
    \small
    \centering
    \vspace{-1mm}
    \resizebox{0.93\textwidth}{!}{
    \begin{tabular}{c | c c c c c c c c c c c c}
    \toprule
      & \multicolumn{12}{c}{CMU-MOSEI Emotion} \\
      Models & \multicolumn{2}{c}{Happy} & \multicolumn{2}{c}{Sad} & \multicolumn{2}{c}{Angry} & \multicolumn{2}{c}{Fear} & \multicolumn{2}{c}{Disgust} & \multicolumn{2}{c}{Surprise}\\
        & Acc & F1 & Acc & F1 & Acc & F1 & Acc & F1 & Acc & F1 & Acc & F1 \\
      \midrule
\multicolumn{13}{c}{Non-Interpretable Methods} \\
\midrule
      EF-LSTM & 68.6 & 68.6 & 74.6 & 70.5 & 76.5 & 72.6 & 90.5 & 86.0 & 82.7 & 80.8 & 91.7 & 87.8 \\
      LF-LSTM & 68.0 & 68.0 & 73.9 & 70.6 & 76.1 & 72.4 & 90.5 & 85.6 & 82.7 & 80.3 & 91.7 & 87.8 \\
      MulT~\cite{tsai2019multimodal} & \textbf{71.8} & \textbf{71.8} & 75.7 & \textbf{73.2} & \textbf{77.6} & \textbf{73.4} & \textbf{90.5} & \textbf{86.0} & \textbf{84.4} & \textbf{83.2} & \textbf{91.7} & \textbf{87.8} \\
      \midrule
\multicolumn{13}{c}{Interpretable Methods} \\
\midrule
      GAM~\cite{hastie2017generalized} & 69.6 & \textbf{69.6} & \textbf{76.2} & 67.8 & 77.5 & 69.7 & \textbf{90.5} & \textbf{86.0} & \textbf{84.2} & 80.7 & \textbf{91.7} & \textbf{87.8} \\
      Multimodal Routing$^*$ & 69.4 & 69.3 & 76.2 & 68.8 & 77.5 & 69.1 & 90.5 & 86.0 & 84.1 & 81.6 & 91.7 & 87.8\\
      Multimodal Routing & \textbf{69.7} & \textbf{69.4} & \textbf{76.0} & \textbf{72.1} & \textbf{77.6} & \textbf{72.8} & \textbf{90.5} & \textbf{86.0} & 83.1 & \textbf{82.3} & \textbf{91.7} & \textbf{87.8}\\
    \bottomrule
       
    \end{tabular}}
    \vspace{-2mm}
    \caption{CMU-MOSEI emotion recognition. Multimodal Routing$^*$ denotes our method without iterative routing. We make our results bold if it is the best or close to the best ($\leq 1\%$).}
    \vspace{-2mm}
    \label{tab:mosei_emo}
\end{table*}

\section{Experiments}
\label{sec:exp}

In this section, we first provide details of experiments we perform and comparison between our proposed model and state-of-the-art (SOTA) method, as well as baseline models. We include interpretability analysis in the next section.

\subsection{Datasets}
We perform experiments on two publicly available benchmarks for human multimodal affect recognition: CMU-MOSEI~\cite{zadeh2018multimodal} and IEMOCAP~\cite{busso2008iemocap}. 
CMU-MOSEI~\cite{zadeh2018multimodal} contains $23,454$ movie review video clips taken from YouTube. For each clip, there are two tasks: sentiment prediction (multiclass classification) and emotion recognition (multilabel classification). For the sentiment prediction task, each sample is labeled by an integer score in the range $[-3, 3]$, indicating highly negative sentiment ($-3$) to highly positive sentiment ($+3$). We use some metrics as in prior work~\cite{zadeh2018multimodal}: seven class accuracy ($Acc_7$: seven class classification in $\mathbb{Z} \in [-3, 3]$), binary accuracy ($Acc_2$: two-class classification in $\{-1, +1\}$), and F1 score of predictions. For the emotion recognition task, each sample is labeled by one or more emotions from \{Happy, Sad, Angry, Fear, Disgust, Surprise\}. We report the metrics~\cite{zadeh2018multimodal}: six-class accuracy (multilabel accuracy of predicting six emotion labels) and F1 score.




IEMOCAP consists of $10$K video clips for human emotion analysis. Each clip is evaluated and then assigned (possibly more than one) labels of emotions, making it a multilabel learning task. Following prior work and insight~\cite{tsai2019multimodal,tripathi2018multi, jack2014dynamic}, we report on four emotions (happy, sad, angry, and neutral), with metrics four-class accuracy and F1 score.

For both datasets, we use the extracted features from a public SDK \url{https://github.com/A2Zadeh/CMU-MultimodalSDK}, whose features are extracted from textual (GloVe word embedding \cite{pennington2014glove}), visual (Facet \cite{imotions}), and acoustic (COVAREP \cite{degottex2014covarep}) modalities. The acoustic and vision features are processed to be aligned with the words (i.e., text features). We present results using this word-aligned setting in this paper, but ours can work on unaligned multimodal language sequences. The train, valid and test set split are following previous work~\cite{wang2019words,tsai2019multimodal}.



\subsection{Ablation Study and Baseline Models}

We provide two ablation studies for interpretable methods as baselines: The first is based on Generalized Additive Model (GAM)~\cite{hastie2017generalized} which directly sums over unimodal, bimodal, and trimodal features and then applies a linear transformation to obtain a prediction. This is equivalent to only using weight $p_i$ and no routing coefficients. The second is our denoted as Multimodal Routing$^*$, which performs only one routing iteration (by setting $t=1$ in Procedure~\ref{algo:routing}) and does not iteratively adjust the routing and update the concepts. 

We also choose other non-interpretable methods that achieved state-of-the-art to compare the performance of our approach to: Early Fusion LSTM (EF-LSTM), Late Fusion LSTM (LF-LSTM)~\cite{hochreiter1997long}, Recurrent Attended Variation Embedding Network (RAVEN)~\cite{wang2019words}, and Multimodal Transformer~\cite{tsai2019multimodal}. 



\subsection{Results and Discussions}

We trained our model on 1 RTX 2080 GPU. We use $7$ layers in the Multimodal Transformer, and choose the batch size as $32$. The model is trained with initial learning rate of $10^{-4}$ and Adam optimizer.


{\bf CMU-MOSEI sentiment.} Table~\ref{tab:mosei_senti} summarizes the results on this dataset. We first compare all the interpretable methods. We see that Multimodal Routing enjoys performance improvement over both GAM~\cite{hastie2017generalized}, a linear model on encoded features, and Multimodal Routing$^*$, a non-iterative feed-forward net with same parameters as Multimodal Routing. The improvement suggests the proposed iterative routing can obtain a more robust prediction by dynamically associating the features and the concepts of the model's predictions. Next, when comparing to the non-interpretable methods, Multimodal Routing outperforms EF-LSTM, LF-LSTM and RAVEN models and performs competitively when compared with MulT~\cite{tsai2019multimodal}. It is good to see that our method can competitive performance with the added advantage of local and global interpretability (see analysis in the later section). The configuration of our model is in the supplementary file.


{\bf CMU-MOSEI emotion.} We report the results in Table~\ref{tab:mosei_emo}. We do not report RAVEN~\cite{wang2019words} and MulT~\cite{tsai2019multimodal} since they did not report CMU-MOSEI results. Compared with all the baselines, Multimodal Routing performs again competitively on most of the results metrics. We note that the distribution of labels is skewed (e.g., there are disproportionately very few samples labeled as ``surprise''). Hence, this skewness somehow results in the fact that all models end up predicting not ``surprise'', thus the same accuracy for ``surprise'' across all different approaches. 


{\bf IEMOCAP emotion.} When looking at the IEMOCAP results in Table~\ref{tab:mosei_senti}, we see similar trends with CMU-MOSEI sentiment and CMU-MOSEI emotion, that multimodal routing achieves performance close to the SOTA method. We see a performance drop in the emotion ``happy'', but our model outperforms the SOTA method for the emotion ``angry''.

\section{Interpretation Analysis}






In this section, we revisit our initial research question: how to locally identify the importance or contribution of unimodal features and the bimodal or trimodal interactions? We provide examples in this section on how multimodal routing can be used to see the variation of contributions. We first present the local interpretation and then the global interpretation.

\begin{figure*}[h!]
    \centering
    \vspace{-2mm}
    \includegraphics[scale=0.075]{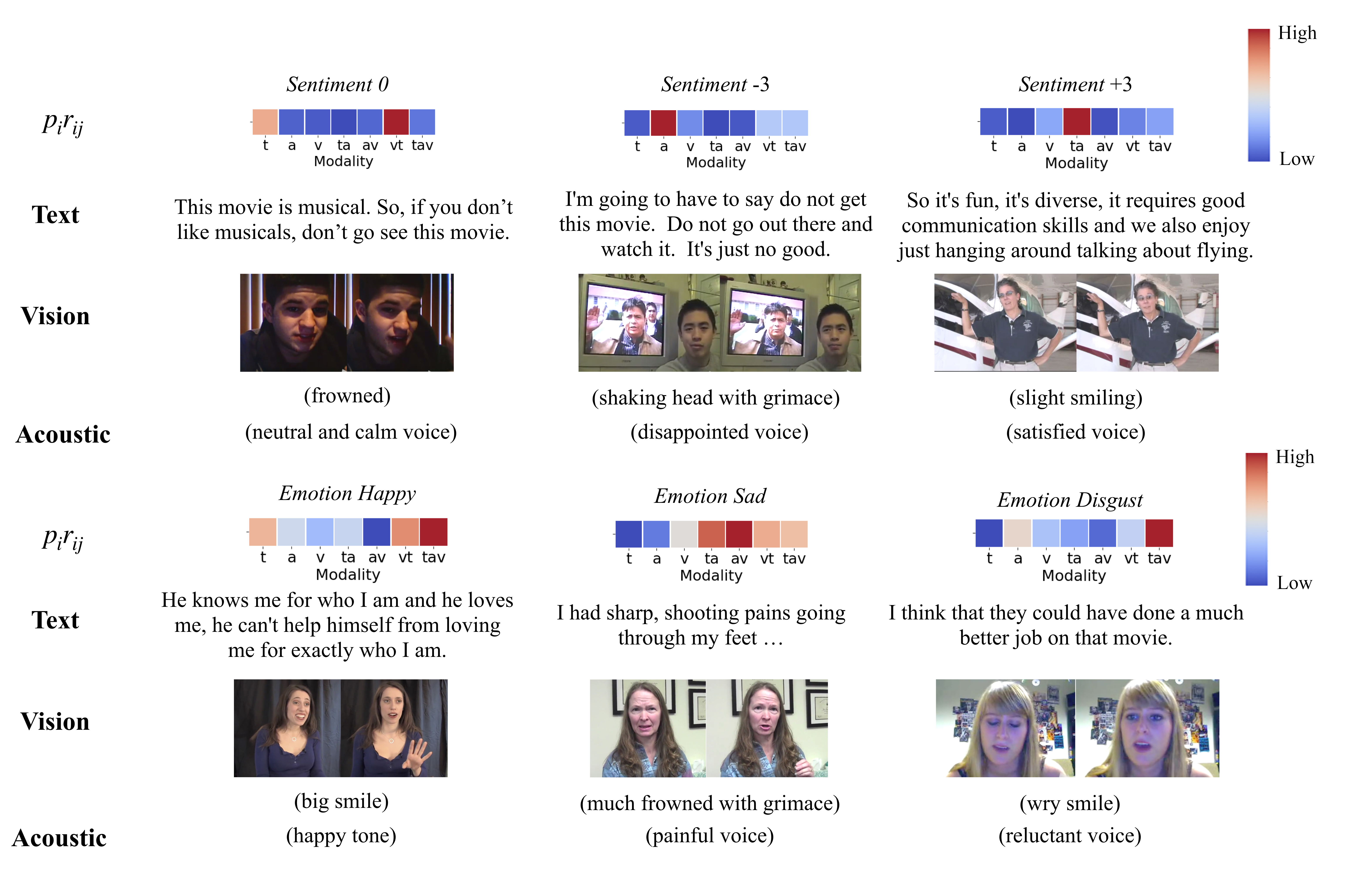}
    \vspace{-11mm}
    \caption{Local interpretation (qualitative results) for Multimodal Routing. The upper row contains three examples from CMU-MOSEI sentiment prediction task; the bottom row contains three examples from CMU-MOSEI emotion recognition task. $p_i r_{ij}$ represents the contribution from the explanatory features $i$ (unimodal/bimodal/trimodal interaction features) to the prediction ${\rm logit}_j$ (see eq.~\ref{eq:logit}). In these examples, $j$ is chosen to be the ground truth label.}
    \label{fig:qualitative}
    \vspace{-2mm}
\end{figure*}

\subsection{Local Interpretation Analysis}

We show how our model makes decisions locally for each specific input sample by looking at inferred coefficients $p_ir_{ij}$. Different samples create different $p_{i}$ and $r_{ij}$, and their product represents how each feature vector contributes to final prediction locally, thus providing local interpretability. We provide such interpretability analysis on examples from CMU-MOSEI sentiment prediction and emotion recognition, and illustrate them in Fig.~\ref{fig:qualitative}. For sentiment prediction, we show samples with true labels neutral (0), most negative sentiment ($-3$), and most positive ($+3$) sentiment score. For emotion recognition, we illustrate examples with true label ``happy'', ``sad'', and ``disgust'' emotions. A color leaning towards red in the rightmost spectrum stands for a high association, while a color leaning towards blue suggests a low association.

In the upper-left example in Fig.~\ref{fig:qualitative}, a speaker is introducing movie \textit{Sweeny Todd}. He says the movie is a musical and suggests those who dislike musicals not to see the movie. Since he has no personal judgment on whether he personally likes or dislikes the movie, his sentiment is classified as neutral (0), although the text modality (i.e., transcript) contains a ``don't". In the vision modality (i.e., videos), he frowns when he mentions this movie is musical, but we cannot conclude his sentiment to be neutral by only looking at the visual modality. By looking at both vision and text together (their interaction), the confidence in neutral is high. The model gives the text-vision interaction feature a high value of $p_ir_{ij}$ to suggest it highly contributes to the prediction, which confirms our reasoning above.

Similarly, for the bottom-left example, the speaker is sharing her experience on how to audition for a Broadway show. She talks about a very detailed and successful experience of herself and describes ``love" in her audition monologue, which is present in the text. Also, she has a dramatic smile and a happy tone. We believe all modalities play a role in the prediction. As a result, the trimodal interaction feature contributes significantly to the prediction of happiness, according to our model.

Notably, by looking at the six examples overall, we could see each individual sample bears a different pattern of feature importance, even when the sentiment is the same. This is a good debuging and interpretation tool. For global interpretation, all these samples will be averaged giving more of a general trend.
\begin{table*}[]
    \centering
    \small
    \resizebox{\textwidth}{!}{%
    \begin{tabular}{cccccccc}
    \toprule
      & \multicolumn{7}{c}{Sentiment} \\
      & -3   & -2 & -1 & 0 & 1 & 2 & 3 \\
      \midrule
      $\overline{r_t}$  & (0.052, 0.066) & \textbf{(0.349, 0.511)} & (0.094, 0.125) & \textbf{(0.194, 0.328)} & (0.078, 0.166) & (0.025, 0.042) & (0.017, 0.039) \\
      $\overline{r_a}$  & \textbf{(0.531, 0.747)} & (0.033, 0.066) & (0.044, 0.075) & (0.051, 0.079) & (0.054, 0.123) & (0.045, 0.069) & (0.025, 0.040) \\
      $\overline{r_v}$  & (0.152, 0.160) & (0.122, 0.140) & (0.205, 0.220) & (0.161, 0.178) & (0.131, 0.140) & (0.128, 0.137) & (0.066, 0.071) \\
      $\overline{r_{ta}}$  & (0.012, 0.030) & (0.012, 0.025) & (0.018, 0.037) & (0.011, 0.045) & (0.014, 0.033) & (0.021, 0.096) & \textbf{(0.728, 0.904)} \\
      $\overline{r_{av}}$  & (0.062, 0.087) & (0.050, 0.064) & \textbf{(0.289, 0.484)} & (0.060, 0.093) & (0.057, 0.079) & (0.153, 0.305) & (0.042, 0.051) \\
      $\overline{r_{vt}}$  & (0.167, 0.181) & (0.174, 0.228) & (0.167, 0.190) & (0.158, 0.172) & (0.122, 0.132) & (0.104, 0.119) & (0.052, 0.062) \\
      $\overline{r_{tav}}$ & (0.112, 0.143) & (0.062, 0.093) & (0.119, 0.149) & (0.149, 0.178) & (0.100, 0.149) & \textbf{(0.213, 0.322)} & (0.064, 0.094)\\
    \bottomrule
    \end{tabular}}
    \vspace{-2mm}
\end{table*}
\begin{table*}[]
    \centering
    \small
    \resizebox{\textwidth}{!}{%
    \begin{tabular}{ccccccc}
    \toprule
      & \multicolumn{6}{c}{Emotions} \\
      & Happy   & Sad & Angry & Fear & Disgust & Surprise \\
      \midrule
      $\overline{r_t}$  & (0.114, 0.171) & (0.078, 0.115) & (0.093, 0.170) & \textbf{(0.382, 0.577)} & (0.099, 0.141) & (0.026, 0.120) \\
      $\overline{r_a}$  & (0.107, 0.171) & (0.095, 0.116) & (0.104, 0.149) & (0.119, 0.160) & \textbf{(0.285, 0.431)} & (0.092, 0.117) \\
      $\overline{r_v}$  & (0.139, 0.164) & (0.143, 0.168) & (0.225, 0.259) & (0.159, 0.182) & (0.141, 0.155) & (0.123, 0.136) \\
      $\overline{r_{ta}}$  & (0.117, 0.158) & (0.039, 0.059) & (0.104, 0.143) & (0.055, 0.079) & (0.055, 0.082) & \textbf{(0.462, 0.615)} \\
      $\overline{r_{av}}$  & (0.102, 0.136) & (0.054, 0.074) & \textbf{(0.358, 0.482)} & (0.219, 0.261) & (0.043, 0.072) & (0.092, 0.107) \\
      $\overline{r_{vt}}$  & (0.173, 0.215) & (0.075, 0.099) & (0.212, 0.241) & (0.180, 0.196) & (0.134, 0.150) & (0.132, 0.142) \\
      $\overline{r_{tav}}$ & (0.182, 0.225) & (0.146, 0.197) & (0.146, 0.183) & (0.158, 0.209) & (0.151, 0.176) & (0.101, 0.116) \\
    \bottomrule
    \end{tabular}}
    \caption{Global interpretation (quantitative results) for Multimodal Routing. Confidence Interval of $\overline{r_{ij}}$, sampled from CMU-MOSEI sentiment task (top) and emotion task (bottom). We bold the values that have the largest mean in each emotion and are significantly larger than a uniform routing ($1/J = 1/7 = 0.143$).}
\label{tab:ci_r}
\end{table*}

\subsection{Global Interpretation Analysis}

Here we analyze the global interpretation of Multimodal Routing. Given the averaged routing coefficients $\overline{r_{ij}}$ generated and aggregated locally from samples, we want to know the overall connection between each modality or modality interaction and each concept across the whole dataset. To evaluate these routing coefficients we will compare them to uniform weighting, i.e., $\frac{1}{J}$ where $J$ is the number of concepts. To perform such analysis,  we provide confidence intervals of each $\overline{r_{ij}}$. If this interval is outside of $\frac{1}{J}$, we can interpret it as a distinguishably significant feature. See Supplementary for similar analysis performed on $\overline{p_ir_{ij}}$ and  $\overline{p_{i}}$.

First we provide confidence intervals of  $\overline{r_{ij}}$ sampled from CMU-MOSEI sentiment. We compare our confidence intervals with the value $\frac{1}{J}$. From top part of Table \ref{tab:ci_r}, we can see that our model relies identified language modality for neutral sentiment predictions; acoustic modality for extremely negative predictions (row $\overline{r_a}$ column -3); and text-acoustic bimodal interaction for extremely positive predictions (row $\overline{r_{ta}}$ column 3). Similarly, we analyze $\overline{r_{ij}}$ sampled from CMU-MOSEI emotion (bottom part of Table \ref{tab:ci_r}). We can see that our model identified the text modality for predicting emotion fear (row $\overline{r_t}$ column \textit{Fear}, the same indexing for later cases), the acoustic modality for predicting emotion disgust, the text-acoustic interaction for predicting emotion surprise, and the acoustic-visual interaction for predicting emotion angry. For emotion happy and sad, either trimodal interaction has the most significant connection, or the routing is not significantly different among modalities. 

Interestingly, these results echo previous research. In both sentiment and emotion cases, acoustic features are crucial for predicting negative sentiment or emotions. This well aligns with research results in behavior science~\cite{lima2013voices}. Furthermore, \cite{livingstone2018ryerson} showed that the intensity of emotion angry is stronger in acoustic-visual than in either acoustic or visual modality in human speech.

\section{Conclusion}
In this paper, we presented Multimodal Routing to identify the contributions from unimodal, bimodal and trimodal explanatory features to predictions in a locally manner. For each specific input, our method dynamically associates an explanatory feature with a prediction if the feature explains the prediction well. Then, we interpret our approach by analyzing the routing coefficients, showing great variation of feature importance in different samples. We also conduct global interpretation over the whole datasets, and show that the acoustic features are crucial for predicting negative sentiment or emotions, and the acoustic-visual interactions are crucial for predicting emotion angry. These observations align with prior work in psychological research. The advantage of both local and global interpretation is achieved without much loss of performance compared to the SOTA methods. We believe that this work sheds light on the advantages of understanding human behaviors from a multimodal perspective, and makes a step towards introducing more interpretable multimodal language models.

\newpage
\section*{Acknowledgements}
This work was supported in part by the DARPA grants FA875018C0150 
HR00111990016, NSF IIS1763562, NSF Awards \#1750439 \#1722822, National Institutes of Health, and Apple.
We would also like to acknowledge NVIDIA’s GPU support.

\bibliographystyle{acl_natbib}
\bibliography{emnlp2020}
\appendix


\begin{table*}[]
    \centering
    \small
    \resizebox{\textwidth}{!}{%
    \begin{tabular}{cccccccc}
    \toprule
      & \multicolumn{7}{c}{Sentiment} \\
      & -3   & -2 & -1 & 0 & 1 & 2 & 3 \\
      \midrule
      $\overline{p_tr_t}$  & (0.060, 0.088) & \textbf{(0.326, 0.538)} & (0.098, 0.141) & (0.128, 0.206) & (0.067, 0.169) & (0.033, 0.060) & (0.010, 0.024) \\
      $\overline{p_ar_a}$  & \textbf{(0.587, 0.789)} & (0.027, 0.055) & (0.035, 0.066) & (0.040, 0.069) & (0.042, 0.105) & (0.027, 0.043) & (0.023, 0.039) \\
      $\overline{p_vr_v}$  & (0.063, 0.082) & (0.060, 0.080) & (0.093, 0.128) & (0.067, 0.089) & (0.057, 0.074) & (0.051, 0.064) & (0.028, 0.038) \\
      $\overline{p_{ta}r_{ta}}$  & (0.015, 0.032) & (0.015, 0.033) & (0.022, 0.045) & (0.037, 0.092) & (0.010, 0.029) & (0.023, 0.113) & \textbf{(0.610, 0.790)} \\
      $\overline{p_{av}r_{av}}$  & (0.060, 0.090) & (0.046, 0.064) & \textbf{(0.227, 0.429)} & (0.030, 0.058) & (0.064, 0.089) & (0.126, 0.258) & (0.038, 0.052) \\
      $\overline{p_{vt}r_{vt}}$  & (0.069, 0.093) & (0.053, 0.104) & (0.080, 0.110) & (0.076, 0.098) & (0.055, 0.073) & (0.049, 0.064) & (0.028, 0.039) \\
      $\overline{p_{tav}r_{tav}}$ & (0.096, 0.119) & (0.056, 0.083) & (0.135, 0.163) & (0.113, 0.164) & (0.080, 0.122) & \textbf{(0.244, 0.394)} & (0.071, 0.133)\\
    \bottomrule
    \end{tabular}}
    \caption{Global interpretation (quantitative results) for Multimodal Routing. Confidence Interval of $\overline{p_ir_{ij}}$, sampled from CMU-MOSEI sentiment task.}
    \label{tab:ci_pr}
\end{table*}


\begin{table*}[]
    \centering
    \small
    \resizebox{\textwidth}{!}{%
    \begin{tabular}{ccccccc}
    \toprule
      & \multicolumn{6}{c}{Emotions} \\
      & Happy & Sad & Angry & Fear & Disgust & Surprise \\
      \midrule
      $\overline{p_tr_t}$  & (0.137, 0.183) & (0.071, 0.099) & (0.107, 0.174) & \textbf{(0.280, 0.481)} & (0.106, 0.138) & (0.068, 0.123) \\
      $\overline{p_ar_a}$  & (0.105, 0.156) & (0.094, 0.113) & (0.104, 0.149) & (0.129, 0.160) & \textbf{(0.310, 0.442)} & (0.078, 0.099) \\
      $\overline{p_vr_v}$  & (0.123, 0.141) & (0.099, 0.129) & (0.189, 0.221) & (0.141, 0.162) & (0.119, 0.128) & (0.103, 0.114) \\
      $\overline{p_{ta}r_{ta}}$  & (0.070, 0.101) & (0.045, 0.065) & (0.127, 0.165) & (0.052, 0.078) & (0.044, 0.065) & \textbf{(0.504, 0.648)} \\
      $\overline{p_{av}r_{av}}$  & (0.104, 0.138) & (0.059, 0.076) & \textbf{(0.286, 0.395)} & (0.200, 0.252) & (0.062, 0.100) & (0.089, 0.102) \\
      $\overline{p_{vt}r_{vt}}$  & (0.131, 0.173) & (0.050, 0.068) & (0.152, 0.199) & (0.122, 0.149) & (0.096, 0.118) & (0.093, 0.115) \\
      $\overline{p_{tav}r_{tav}}$ & (0.160, 0.187) & (0.132, 0.197) & (0.132, 0.174) & (0.151, 0.183) & (0.151, 0.173) & (0.096, 0.111) \\
    \bottomrule
    \end{tabular}}
    \caption{Global interpretation (quantitative results) for Multimodal Routing. Confidence Interval of $\overline{p_ir_{ij}}$, sampled from CMU-MOSEI emotion task.}
    \label{tab:ci_pr_emo}
\end{table*}

\begin{table}[H]
\small
\centering
\resizebox{.25\textwidth}{!}{
\begin{tabular}{cc}
\toprule
 & Confidence Interval \\
\midrule
$\overline{p_t}$  & (0.98, 0.995) \\
$\overline{p_a}$  & (0.991, 0.992) \\
$\overline{p_v}$  & (0.807, 0.880) \\
$\overline{p_{ta}}$  & (0.948, 0.965) \\
$\overline{p_{av}}$  & (0.968, 0.969) \\
$\overline{p_{vt}}$  & (0.588, 0.764) \\
$\overline{p_{tav}}$ & (0.908, 0.949) \\
\bottomrule
\end{tabular}}
    \caption{Global interpretation (quantitative results) for Multimodal Routing. Confidence interval of $\overline{p_i}$, sampled from CMU-MOSEI sentiment task.}
    \label{tab:ci_aa_senti}
\end{table}


\begin{table}[H]
\tiny
\centering
\resizebox{.3\textwidth}{!}{
\begin{tabular}{cc}
\toprule
 & Confidence Interval \\
\midrule
$\overline{p_t}$  & (0.980, 0.999) \\
$\overline{p_a}$  & (0.991, 0.992) \\
$\overline{p_v}$  & (0.816, 0.894) \\
$\overline{p_{ta}}$  & (0.935, 0.963) \\
$\overline{p_{av}}$  & (0.967, 0.968) \\
$\overline{p_{vt}}$  & (0.635, 0.771) \\
$\overline{p_{tav}}$ & (0.913, 0.946) \\
\bottomrule
\end{tabular}}
    \caption{Global interpretation (quantitative results) for Multimodal Routing. Confidence interval of $\overline{p_i}$, sampled from CMU-MOSEI emotion task.}
    \label{tab:ci_aa_emo}
\end{table}

\subsection{Encoding $p_i$ from input}
In practice, we use the same MulT to encode $f_i$ and $p_i$ simultaneously. We design MulT to have an output dimension $d_f + 1$. A sigmoid function is applied to the last dimension of the output. For this output, the first $d_f$ dimensions refers to $f_i$ and the last dimension refers to $p_i$.

\subsection{Training Details and Hyper-parameters}

Our model is trained using the Adam~\cite{kingma2014adam} optimizer with a batch size of 32. The learning rate is 1e-4 for CMU-MOSEI Sentiment and IEMOCAP, and 1e-5 for CMU-MOSEI emotion. We apply a dropout~\cite{srivastava2014dropout} of 0.5 during training. 

For the encoding stage, we use MulT~\cite{tsai2019multimodal} as feature extractor. After the encoder producing unimodal, bimodal, and trimodal features, we performs linear transformation for each feature, and output feature vectors with dimension $d_f = 64.$  

We perform two iterations of routing between features and concepts with dimension $d_c = 64$ where $d_c$ is the dimension of concepts. All experiments use the same hyper-parameter configuration in this paper.
\subsection{Remarks on CMU-MOSEI Sentiment}
Our model poses the problem as classification and predicts only integer labels, so we don't provide mean average error and correlation metrics.

\subsection{Remarks on CMU-MOSEI Emotion}
Due to the introduction of concepts in our model, we transform the CMU-MOSEI emotion recognition task from a regression problem (every emotion has a score in $[0, 3]$ indicating how strong the evidence of that emotion is) to a classification problem. For each sample with six emotion scores, we label all emotions with scores greater than zero to be present in the sample. Then a data sample would have a multiclass label.



\subsection{Global Interpretation Result}
We analyze global interpretation of both CMU-MOSEI sentiment and emotion task.

\paragraph{CMU-MOSEI Sentiment} The analysis of the routing coefficients $\overline{r_{ij}}$ is included in the main paper. We then analyze $\overline{p_i}$ (table \ref{tab:ci_aa_senti}) and the products $\overline{p_ir_{ij}}$ (table \ref{tab:ci_pr}). Same as analysis in the main paper, our model relies on acoustic modality for extremely negative predictions (row $\overline{r_a}$ column -3) and text-acoustic bimodal interaction for extremely positive predictions (row $\overline{r_{ta}}$ column 3). The sentiment that is neutral or less extreme are predicted by contributions from many different modalities / interactions. The activation table shows high activation value ($ >  0.8$) for most modality / interactions except $\overline{p_{vl}}$.

\paragraph{CMU-MOSEI Emotion}Same as above, we analyze $\overline{p_i}$ (Table \ref{tab:ci_aa_emo}) and the product $\overline{p_ir_{ij}}$ (Table \ref{tab:ci_pr_emo}).The result is very similar to that of $\overline{r_{ij}}$. The activation table shows high activation value ($ >  0.8$) for most modality / interactions except $\overline{p_{vl}}$, same as CMU-MOSEI sentiment. We see strong connections between audio-visual interactions and angry, text modality and fear, audio modality and disgust, and text-audio interactions and surprise. The activation table shows high activation value ($ >  0.8$) for most modality / interactions except $\overline{p_{vl}}$ as well.


\label{linearform}

\end{document}